%% file: root.tex
\documentclass[letterpaper, 10 pt, conference]{ieeeconf}
\IEEEoverridecommandlockouts
\usepackage{cite}
\usepackage{amsmath,amssymb,amsfonts}
\usepackage{algorithmic}
\usepackage{graphicx}
\usepackage{textcomp}
\usepackage{xcolor}
\usepackage{subcaption}
\usepackage{multirow}
\usepackage{url}
\usepackage{hyperref}

\begin{document}
\include{commands}

\title{\LARGE \bf 
\our: Enhancing Standard Definition Maps \\ by Incorporating Road Knowledge using LLMs}

\author{Hitvarth Diwanji$^{*1}$,  Jing-Yan Liao$^{*1}$, Akshar Tumu$^{*1}$, \\ Henrik I. Christensen$^{2}$, Marcell Vazquez-Chanlatte$^{3}$, and Chikao Tsuchiya$^{3}$
\thanks{This research is supported by Nissan.}%
\thanks{*These authors contributed equally to this work.}%
\thanks{$^{1}$Contextual Robotics Institute,
University of California San Diego, La Jolla, CA 92093, USA 
{\tt\small \{hdwanji, j3liao, atumu\}@ucsd.edu}}%
\thanks{$^{2}$Dept. of Comp. Sci. and Eng., UC San Diego, La Jolla, CA 92093, USA {\tt\small hichristensen@ucsd.edu}}%
\thanks{$^{3}$Nissan North America, 3400 Central Expy, Santa Clara, CA 95051}%
}

\maketitle

\begin{abstract}
High-definition maps (HD maps) are detailed and informative maps capturing lane centerlines and road elements. Although very useful for autonomous driving, HD maps are costly to build and maintain. Furthermore, access to these high-quality maps is usually limited to the firms that build them. On the other hand,
standard definition (SD) maps provide road centerlines with an accuracy of a few meters. In this paper, we explore the possibility of enhancing SD maps by incorporating information from road manuals using LLMs. We develop \our, an end-to-end pipeline to enhance SD maps with location-dependent road information obtained from a road manual. Our pipeline requires no sensor data input and only relies on road manuals and SD maps. We experiment several ways of using LLMs for map enhancement. Furthermore, we demonstrate the generalization ability of \our\ by showing results from six states in the United States and Japan. Code is available at \href{https://github.com/AutonomousVehicleLaboratory/SDplusplus}{https://github.com/AutonomousVehicleLaboratory/SDplusplus}  
\end{abstract}


\section{Introduction}

High-Definition (HD) maps play a critical role in autonomous driving systems,
supporting key components such as localization, prediction, and planning through precise lane-level detail. HD maps leverage advances in sensor fusion technology, incorporating multiple
modalities that complement each other to provide rich road context. 

Despite their importance, the creation and maintenance of HD maps remain resource-intensive and challenging. Producing these maps requires extensive data
collection using vehicles equipped with high-end sensors like LiDAR. This raw data then undergoes sophisticated processing to extract detailed semantic features like lane boundaries. However, the process often involves significant manual annotation, which is both time-consuming and
labor-intensive. Furthermore, HD maps require frequent updates to accommodate
changes such as construction or layout modifications, making
them costly and difficult to scale. Prior studies~\cite{osmvshdmap} highlight
these operational challenges, emphasizing the need for more efficient and
scalable solutions. To address these limitations, researchers have explored
alternative approaches, such as generating HD maps directly from onboard sensor
data~\cite{li2021hdmapnet, MapTR, li2023toponet, ranganatha2024semvecnetgeneralizablevectormap}. While promising, these methods often face challenges with generalization, limited range, and still require significant data collection.

\begin{figure}
    \centering
    \includegraphics[width=0.95\linewidth]{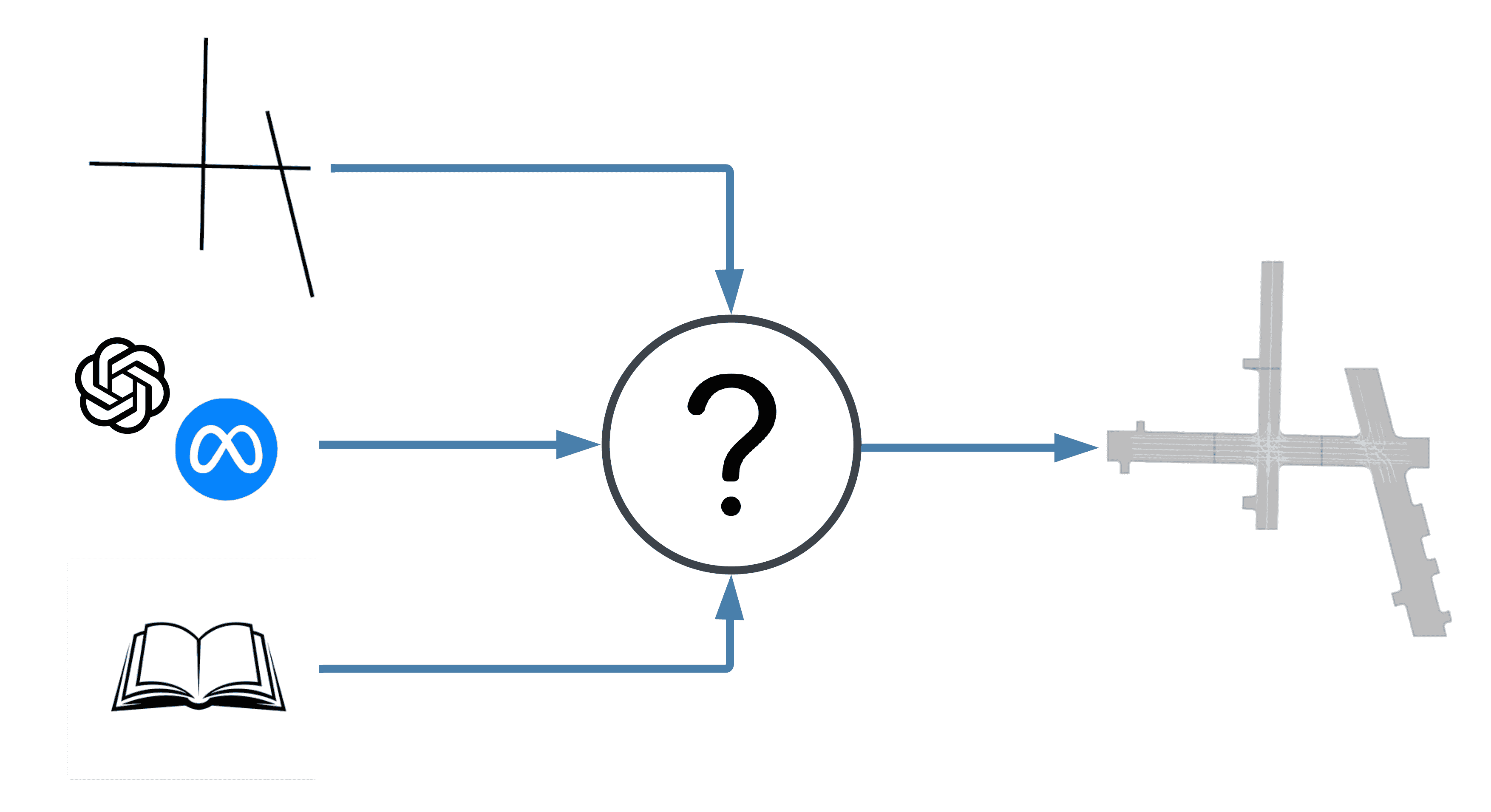}
    \caption{Overview of our proposed pipeline, which explores the potential of leveraging OpenStreetMap, LLMs, and official documents for mapping.}
    \label{fig:teaser}
\end{figure}

Standard-Definition (SD) maps offer a cost-effective
and scalable alternative. OpenStreetMap (OSM)~\cite{OpenStreetMap} represent road networks using basic geographical
coordinates, making them easy to scale. However, SD maps lack the fine-grained
detail and accuracy~\cite{osm_quality} required for many autonomous driving applications, limiting
their direct usability. Recent work~\cite{smerf, ory_iros2024} has attempted to generate HD maps by integrating them with sensor data, improving their usability for lane topology reasoning.  Nevertheless, these methods still face challenges related to generalization and data dependency. Given that most existing approaches extract lane-level details from visual cues, we explore the possibility of improving SD maps using other public resources to help close the gap between SD and HD maps.

In this study, we propose a novel approach to bridge the gap between HD and SD
maps without relying on any sensor data or manual annotation. Specifically, we
explore the use of large language models (LLMs) to enhance SD maps using
publicly available resources, such as highway design manuals (HDM)~\cite{hdm} that
provide detailed road design specifications. By leveraging these resources alongside OSM data, our method aims to generate enriched map representations that approximate the detail and utility of HD maps. Our key contributions are as follows:

\begin{itemize}
\item \textbf{Sensor-Free SD Map Enhancement: }We present a novel method for
  enhancing SD maps without using any sensor data, leveraging open-source
  information and LLMs to automate the process.
\item \textbf{Generalization Across Regions: }We demonstrate the generalization
  ability of our approach by incorporating road design guidelines from multiple
  regions, including various U.S. states and Japan.
\end{itemize}

This work aims to provide a scalable, cost-effective, and informative map prior while retaining its critical functionality for
autonomous driving systems.

\section{Related Work} \label{related_work}
\subsection{High Definition (HD) Maps Generation}
HD map generation has been extensively studied in the context of autonomous
driving, with research broadly categorized into online and offline methods. Both
approaches aim to create detailed, accurate maps but differ significantly in
their operational processes and use cases. Online HD map
generations~\cite{li2021hdmapnet,li2023toponet,MapTR,ranganatha2024semvecnetgeneralizablevectormap, ory_iros2024} estimate locations of map elements based on sensor inputs to
avoid continuous map maintenance. Offline HD map generation methods allow the
aggregation of more information. For example, Zhou et
al.\cite{Zhou2021IROS_HDmap} automate the HD Map building pipeline with instance
segmentation, mapping, and particle filter-based lane aggregation.

\subsection{LLM for Autonomous Driving}
Recently, the autonomous driving domain saw many diverse applications of LLMs.
DriveGPT4~\cite{drivegpt4} and LMDrive~\cite{lmdrive} attempt to approach
autonomous driving in an end-to-end fashion. Elhafsi et al.
\cite{anomalydetection} detect visual anomalies using LLMs to prevent certain
failure modes in autonomous driving. Furthermore, we see numerous works focusing
on either of perception~\cite{hilmd}, prediction~\cite{llm_mp, mtd_gpt, lcllm},
or planning and control~\cite{planagent, dme_driver, agent_driver} aspects of
autonomous driving. However, to the best of our knowledge, we have not found any
work that leverages LLMs for mapping. The closest work we see are around map
annotations for more awareness of the surroundings. Talk2BEV~\cite{talk2bev}
annotates an instantaneous Bird's Eye View (BEV) map with natural language
descriptions of identified map elements (like cars, bikes, etc) using LLMs.
However, their language-enhanced map is frame-dependent, which implies its
single usage. \our\ on the other hand, is a static map and can be reused for
downstream tasks. Moreover, our method focuses on enhancing existing SD maps
without using any sensors.

\section{Prerequisites}
\subsection{OSM Data Format}
An OSM consists of \textit{nodes}, \textit{ways} and \textit{relations}
supplemented with metadata for each data structure. Every node $n=(id, lat,
lon)$ in an OSM has a unique node identifier and represents a geographical
point denoted by latitude and longitude. All the roads are defined using ways. A
way $w = (n_0, n_1, ...)$ is represented as a sequence of nodes that make up the
way. Structures like buildings can also be represented as ways. Finally,
relations are used to indicate logical associations between nodes and ways. In
our case, relations are not used as our focus is to enhance lane-level details
of road elements. 

\subsection{Road Manuals}
\begin{figure}
    \centering
    \includegraphics[width=0.95\linewidth]{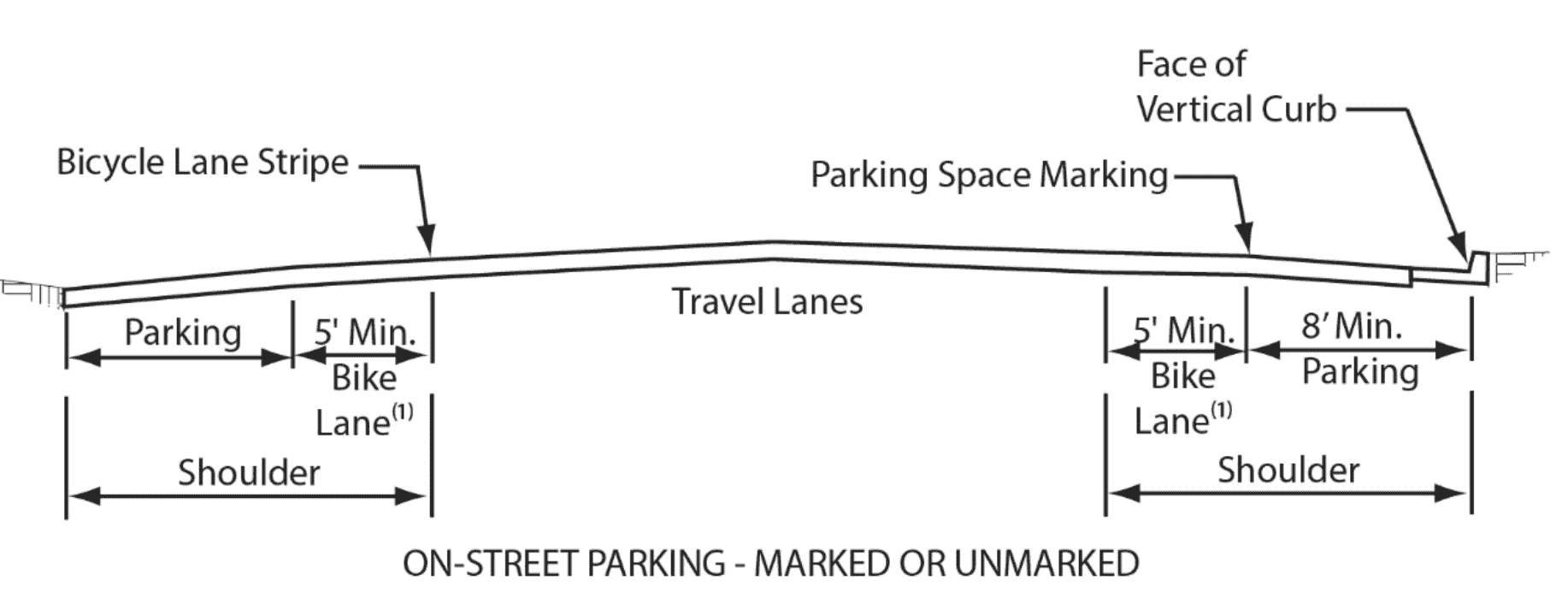}
    \caption{An example illustration from the HDM}
    \label{fig:hdm}
\end{figure}

Highway Design Manual (HDM)~\cite{hdm} is a comprehensive document that provides
standardized guidelines and specifications for the design, construction, and
maintenance of roadways and related infrastructure. It provides detailed
parameters such as lane widths, shoulder dimensions, road alignments, and
traffic control elements like signage and markings, shown in Fig. \ref{fig:hdm}. These manuals are created
and maintained by government transportation departments, such as the U.S.
Department of Transportation (DOT) or state-level agencies to ensure safety,
efficiency, and consistency in highway design across different regions. The contained information is also regional, e.g. in the US we can find HDMs at national, state and city level.

The HDM is particularly valuable for SD map enhancement, as it defines precise
geometric and structural details that can serve as priors for enhancing map
accuracy. For example, the manual specifies road lane configurations, minimum
and maximum widths, and intersection designs, all critical inputs when
augmenting SD maps with lane-level information. With this official document in
hand, \our\ makes our map representation closer to an HD map at a low cost.

\begin{figure*}
\centering \includegraphics[width=0.9\textwidth]{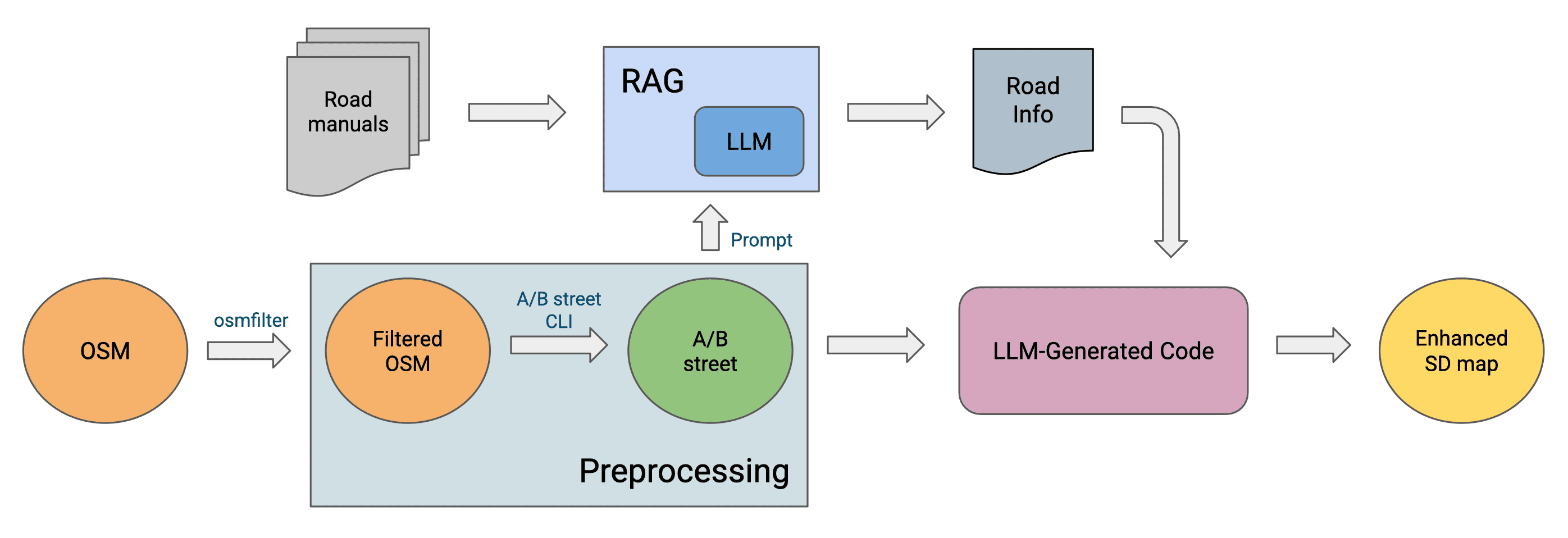}
\caption{\our\ takes OSM as input, filters it by removing non-road elements,
  processes it using A/B street software and prompts the RAG pipeline with the
  basic road information from A/B street. The RAG pipeline maintains a vector
  database of information from road manuals. When prompted with a query, it uses
  an LLM to obtain detailed road information of the queried roads using the
  context from road manuals. This detailed road information then goes to an
  LLM-generated code that gives us the enhanced HD map.}
\label{fig:pipeline}
\end{figure*}

\subsection{Retrieval-Augmented Generation (RAG)}
RAG combines external information retrieval with language model generation. Large documents are split into chunks, embedded via an embedding model, and stored in a vector database. At inference, a user query is embedded and compared with stored embeddings (e.g., using cosine similarity) to retrieve the top-$k$ most relevant chunks. These are appended to the query and fed into the language model to produce a response. This approach allows the model to incorporate domain-specific knowledge, improving accuracy and relevance.

RAG is well-suited for SD map enhancement for several reasons:

\begin{itemize} \item \textbf{Domain Knowledge Integration: }RAG can retrieve relevant details from sources like highway design manuals, helping ensure outputs reflect real-world standards. \item \textbf{Improved Consistency: }By grounding outputs in structured, authoritative data, RAG reduces inconsistencies in map generation. \item \textbf{Adaptability: }RAG enables access to updated information without retraining, supporting dynamic map updates. \item \textbf{Overcoming LLM Limitations: }It enhances the LLM’s ability to handle fine-grained, domain-specific tasks by providing relevant context. \end{itemize}

\section{Methodology}
In this section, we introduce two pipelines for generating enhanced Standard
Definition (SD) maps: Direct Generation and Knowledge-Based Algorithmic
Generation. The comparison of these methods and their results are presented in
Section \ref{experiments}. Inspired by the recent advancements in Large Language
Models (LLMs) for text generation \cite{gpt3, llama}, we explore their potential
to enhance SD maps, given that SD maps, such as OpenStreetMap (OSM) data, are
represented in text form. The Direct Generation method uses a straightforward
approach by combining OSM input, LLM capabilities, and HDM knowledge. In contrast, Knowledge-Based Algorithmic Generation
incorporates Retrieval-Augmented Generation (RAG) techniques and additional
post-processing steps, leveraging LLMs for specific tasks while addressing the
limitations of direct text-based map generation.

\subsection{Direct Generation}

The Direct Generation method tests the capability of LLMs to enhance SD maps
through structured prompting and contextual input.
\subsubsection{Process}
\begin{itemize}
  \item \textbf{Preprocessing OSM Data: }OSM data is first filtered to retain
    only road-related information. This preprocessing step ensures that
    irrelevant details do not overwhelm the LLM.
  \item \textbf{Providing Context with Pre-Prompt: }A carefully designed
    pre-prompt describes the OSM data structure, its common tags, and how these
    relate to HD maps. This step helps the LLM understand the domain-specific
    context and prepares it for processing the input.
  \item \textbf{Incorporating HDM: }The HDM is uploaded
    with instructions for the LLM to extract key road-related details such as
    lane width, shoulder widths, and other essential parameters for SD map
    enhancement.
  \item \textbf{Constraints for HD Map Generation: }The LLM is tasked with
    generating the enhanced map while adhering to specific constraints:
  \begin{enumerate}
      \item All road nodes in the OSM data must be included.
      \item No modifications are allowed to the existing road nodes in OSM.
      \item Any assumptions made during the process must be explicitly stated.
  \end{enumerate}
  \item \textbf{Output Format Specification: }The final output is formatted to
    allow further analysis and validation.
\end{itemize}

\subsubsection{Challenges Identified}

While this method leverages LLMs for generating enhanced maps, its performance
is limited by inconsistencies in text-based map generation, as demonstrated in
Section \ref{experiments}. These limitations led us to design an improved
method: Knowledge-Based Algorithmic Generation.

\subsection{Knowledge-based Algorithmic Generation}
Recognizing that text-based map generation is not a strength of LLMs; we shifted
their role to tasks they excel at, such as extracting structured knowledge from
the HDM. This refined pipeline, shown in Fig.~\ref{fig:pipeline}, integrates
additional preprocessing and post-processing steps to address the shortcomings
of the Direct Generation approach.
\begin{itemize}
  \item \textbf{Enhanced Preprocessing Using A/B Street: }To address
    inconsistencies in OSM data due to its open-source nature, we use the A/B
    Street tool \cite{abstreet} to further refine and standardize road segment
    attributes. This step ensures consistent labels across all road segments.
    Furthermore, it converts the XML format of OSM to JSON format, which is
    easier to parse for an LLM.
  \item \textbf{Segment-Level Information Extraction: }Essential details for
    each road segment—such as lane width, bike lane width, and total road
    width—are extracted from the HDM using LLMs. The LLM focuses solely on
    understanding and retrieving relevant information, leveraging its strengths
    in contextual text analysis.
  \item \textbf{Map Generation: }With the extracted road information, the
    enhanced map is generated through algorithmic methods. These methods ensure
    that the final output adheres to the constraints of HD maps and aligns with
    the HDM specifications.
\end{itemize}

We present three variants of Algorithmic Generation:

\subsubsection{One-Shot Generation (\oneshot)}
Here the complete road information is generated in one run of the pipeline
(shown in Fig. \ref{fig:pipeline}).

\subsubsection{Iterative Generation (\increm)} In this variant, we request the LLM to generate the road information one lane at a time. This is to put more attention to every lane.

\subsubsection{Autoregressive Generation (\autoreg)} Just like \increm\ variant, \autoreg\ generates lanes one-by-one. But here, the pipeline is run in an autoregressive manner. That is, once we generate information for one lane, it is appended to the context for generating the next lane. Hence, the generated lanes benefit from the extracted information of other lanes. 

In terms of cost, \autoreg\ $>$ \increm\ $\gg$ \oneshot. We compare these variants in section \ref{experiments} to see if the added expense is worth the performance gain.


\section{Experiments} \label{experiments}
In this section, we first introduce our implementation details in
\ref{implementation}, and show our qualitative and quantitative results for both
of our approaches in \ref{ourmethod}. Finally, we compare on a subset of
Argoverse 2 dataset~\cite{Argoverse2}, and Tokyo
Japan to demonstrate our generalization ability.
\subsection{Implementational Details}\label{implementation}
For preprocessing, we use osmium~\cite{osmium} and osmfilter~\cite{osmfilter} to
extract and filter OSM data.

In the RAG pipeline, we utilize LangChain~\cite{langchain}. We use chapter 300 of the Highway Design Manual (HDM)~\cite{hdm} as our road manual. The document is
divided into smaller chunks of text using PyPDFLoader~\cite{PyPDFLoader}, and we generate text
embeddings for these chunks using the OpenAI embedding model.

For the generation component of the RAG pipeline, we compare the performances of two famous Large Language Models (LLMs) - GPT-4o and Llama. We select these two models because while the former is proprietary, the latter is open-sourced.

To parse the extracted road information, we use Python code generated by an LLM. This code converts the data into a JSON format, representing sequences of points for roads, lanes, and bike lanes.

\subsection{Direct generation vs Algorithmic Generation}\label{ourmethod}

\begin{figure}[]
\centering
\begin{subfigure}{.5\linewidth}
  \centering
  \includegraphics[width=.95\linewidth]{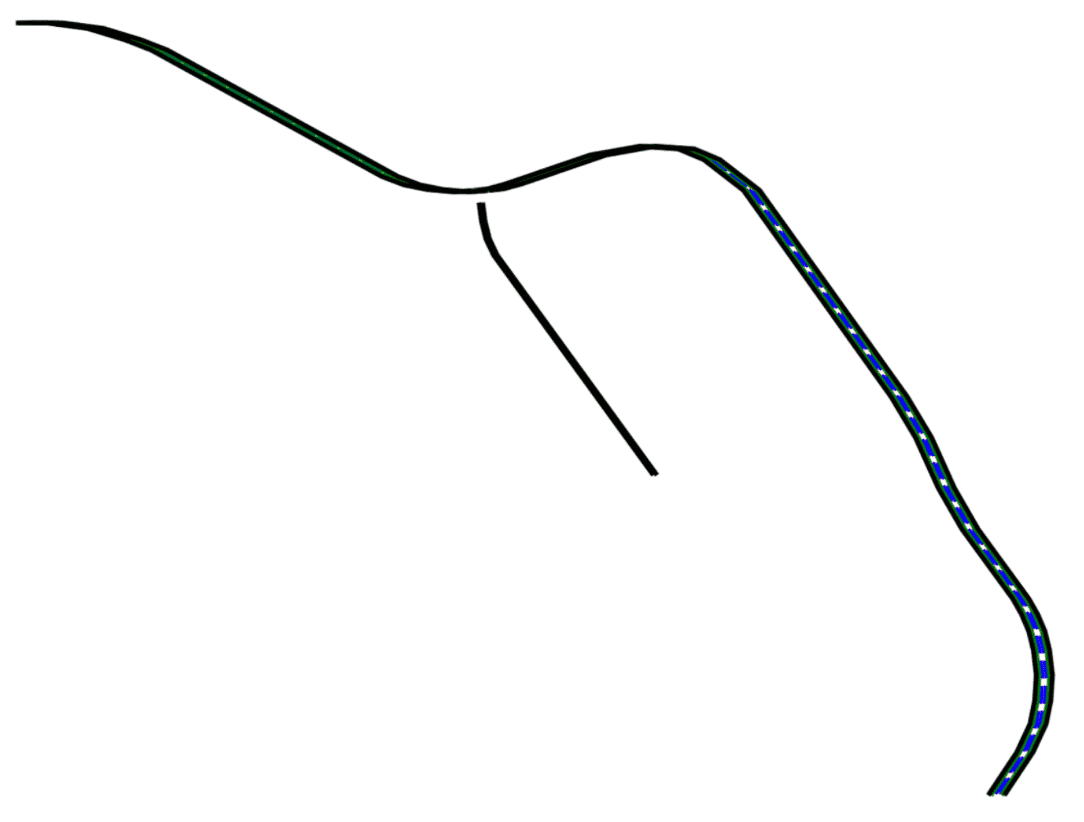}
  \caption{Direct Generation}
\end{subfigure}%
\begin{subfigure}{.5\linewidth}
  \centering
  \includegraphics[width=.95\linewidth]{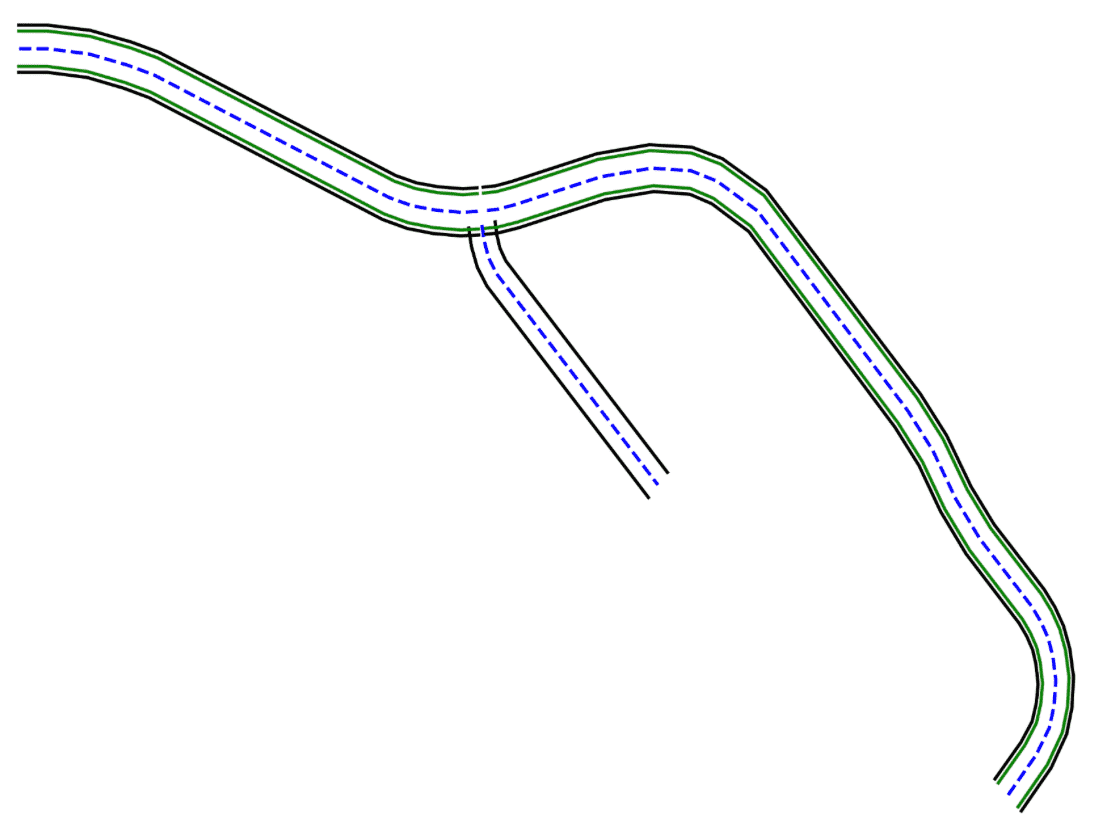}
  \caption{Algorithmic Generation}
\end{subfigure}
\caption{A qualitative example for direct generation vs algorithmic generation}
\label{fig:direct_vs_algo}
\end{figure}

In this experiment, we analyze and compare the outputs of direct generation with
algorithmic generation for a small OSM map spanning about 100 meters. Fig. \ref{fig:direct_vs_algo} 
shows example output for both methods. The lane widths are consistent for algorithmic 
generation but vary noticeably across the directly generated map. The direct generation 
method also fails to comprehend the scale of lane width compared to the map's scale. On 
conducting several runs, we observe that the results of algorithmic
generation are consistent, whereas the direct generation fails to produce clean
outputs. The example shown in Fig. \ref{fig:direct_vs_algo} was the best
of 5 runs. Furthermore, there were cases in which direct generation produced
broken outputs or failed to provide an output altogether.

\begin{table*}[h]
    \centering
    \begin{tabular}{|c|c|c|c|c|c|}
    \hline
         \multirow{2}{*}{Model} & \multirow{2}{*}{Variant} & \multirow{2}{*}{\% of Valid Maps}  & \multirow{2}{*}{Chamfer$_{avg}$ (m)} & \multirow{2}{*}{Chamfer$_{min}$ (m)} & \multirow{2}{*}{Recall} \\ 
          &  &  &  &  &  \\ \hline
         Baseline & N/A & 100 & $3.53 \pm 2.37$ & $0.08$ & $0.73$ \\
         \hline
         GPT-4o ($P1$) & \oneshot & 100 & $24.73 \pm 36.62$ & $0.14$ & $0.39$ \\
         GPT-4o ($P2$)& \oneshot & 100 & $2.52 \pm 2.45$ & $0.1$ & $0.8$ \\
         GPT-4o & \increm & 100 & $2.54 \pm 2.45$ & $0.22$ & $0.8$ \\
         GPT-4o & \autoreg & 100 & $2.91 \pm 2.46$ & $0.08$ & $0.78$ \\ \hline
         Llama ($P1$) & \oneshot & 16 & $4.94 \pm 8.09$ & $0.77$ & $0.01$ \\
         Llama ($P2$) & \oneshot & 65 & $2.65 \pm 2.13$ & $0.38$ & $0.18$ \\
         Llama & \increm  & 100 & $2.5 \pm 2.45$ & $0.1$ & $0.81$ \\
         Llama & \autoreg & 100 & $2.5 \pm 2.45$ & $0.1$ & $0.81$ \\ \hline
    \end{tabular}
    \caption{Quantitative results on Argoverse 2 comparing various variants of \our.}
    \label{tab:av2}
\end{table*}

\subsection{Results on Argoverse 2}
\subsubsection{Evaluation method} We compare the output of \our\ with Argoverse ground truth using two metrics: Averaged Chamfer distance~\cite{chamfer}, and Recall. The reason for using Recall instead of precision is \our\ include larger area than Argoverse ground truth. To calculate the Chamfer distance-based metric, the corresponding road in \our\ map for every road in Argoverse is established by matching their OSM way ID. The correspondence between predicted and ground-truth lanes is determined by minimizing the Chamfer distance. Table \ref{tab:av2} shows the average of all these minimum distances. A \our\ lane prediction is considered correct if its Chamfer Distance to a lane in Argoverse 2 is less than 5 meters, as this threshold roughly corresponds to the typical road width. Following that definition of a correct lane, the recall is calculated as 
\begin{align*}
    recall &= \frac{\text{correct lanes}}{\text{total ground truth lanes}}
\end{align*}

\begin{figure}[]
\centering
\begin{subfigure}{.8\linewidth}
  \centering
  \includegraphics[width=.95\linewidth]{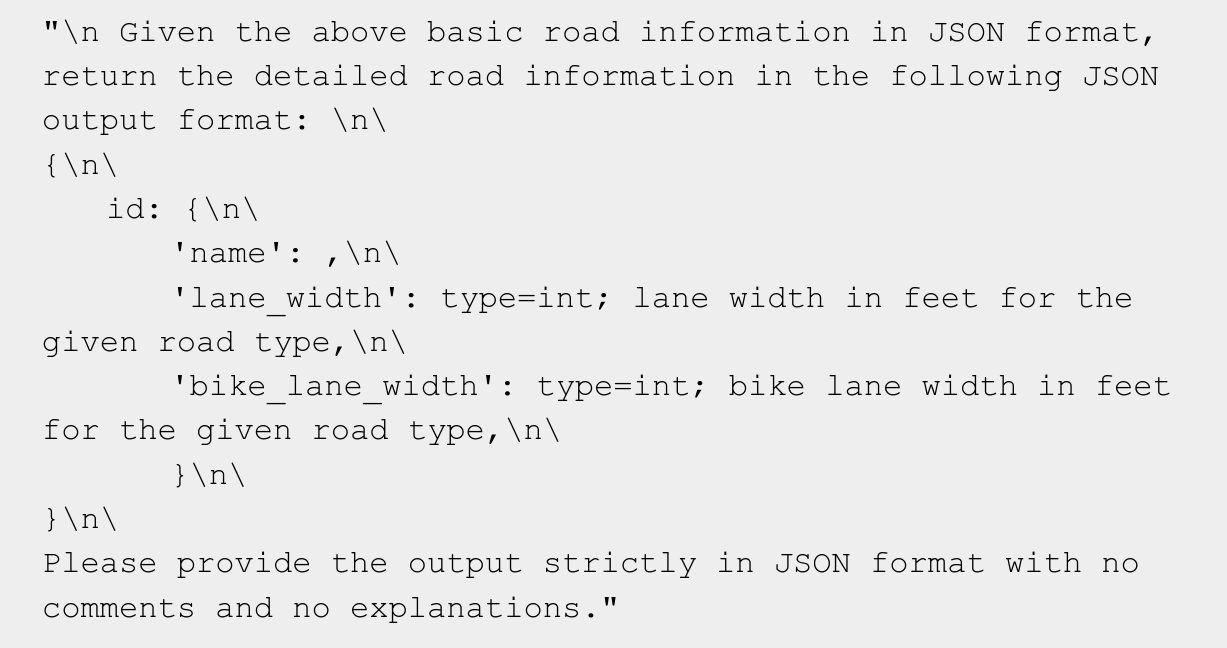}
  \caption{Prompt $P1$}
\end{subfigure}%
\vspace{1.5mm}
\begin{subfigure}{.8\linewidth}
  \centering
  \includegraphics[width=.95\linewidth]{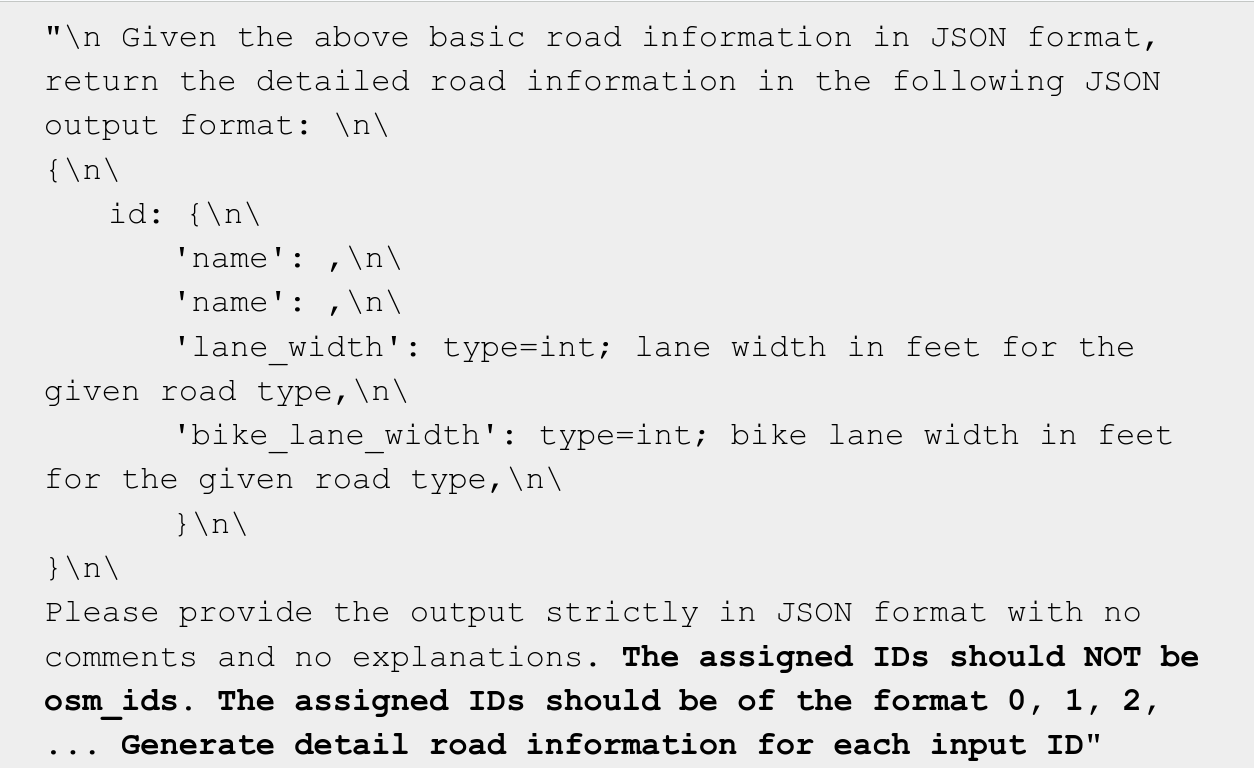}
  \caption{Prompt $P2$}
\end{subfigure}
\caption{Prompts $P1$ (a) and $P2$ (b) used in the RAG pipeline to obtain the detailed road information from the LLM}
\label{fig:prompt}
\end{figure}

\subsubsection{Quantitative results}\label{quantitative} The quantitative evaluation of our approach on Argoverse is provided 2 in Table \ref{tab:av2}. Our experiments reveal several key insights regarding the performance of different language models and the impact of prompt design. The baseline method, A/B Street, uses a hand-crafted, rule-based approach to augment OSM. Interestingly, it performs quite well, which we attribute to the strong domain knowledge embedded by its designer.

First, we compare the performance of two large language models (LLMs), GPT-4o and Llama 3.3. Our results show that GPT-4o consistently outperforms Llama across various methods. Notably, GPT-4o achieves strong results simply by using an improved prompt, whereas Llama benefits more from an iterative generation approach. This difference in performance may be attributed to the model size, as we use the 7-billion-parameter version of Llama, which is significantly smaller than GPT-4o.

Second, we analyze the role of prompt design in improving map generation. Previous studies have shown that even small changes in prompt phrasing can lead to significant variations in LLM outputs~\cite{promptprogramming, calibrateuse}. We observe a similar effect in our \our\ predictions.  As shown in Fig. \ref{fig:prompt}, we compare two prompts, $P1$ and $P2$, and find that the phrasing of the prompt has a notable impact on the quality of generated maps. These findings underscore the importance of both model selection and prompt engineering in improving SD map generation using LLMs. Furthermore, our results demonstrate consistency, as supported by the Chamfer Distance and variance metrics. Since road width is typically around 5 meters, our generated maps could serve as a strong prior, incorporating road connectivity and lane-level information.

\begin{figure}[]
\centering
\begin{subfigure}{.5\linewidth}
  \centering
  \includegraphics[width=.95\linewidth]{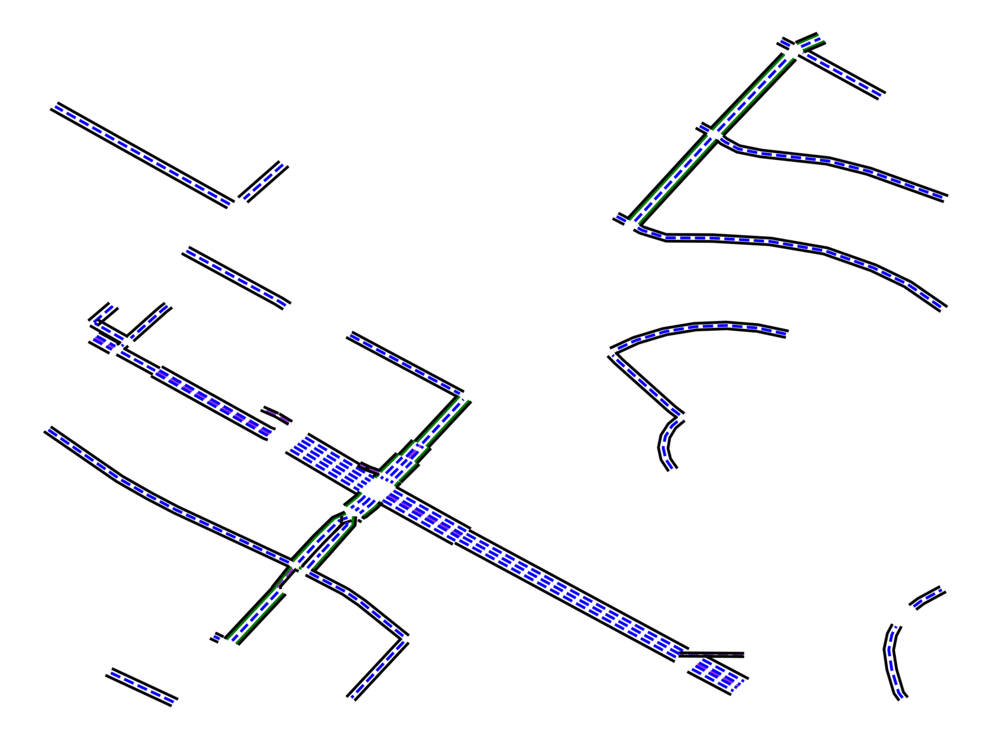}
  \caption{GPT-4o OSG $P1$}
\end{subfigure}%
\begin{subfigure}{.5\linewidth}
  \centering
  \includegraphics[width=.95\linewidth]{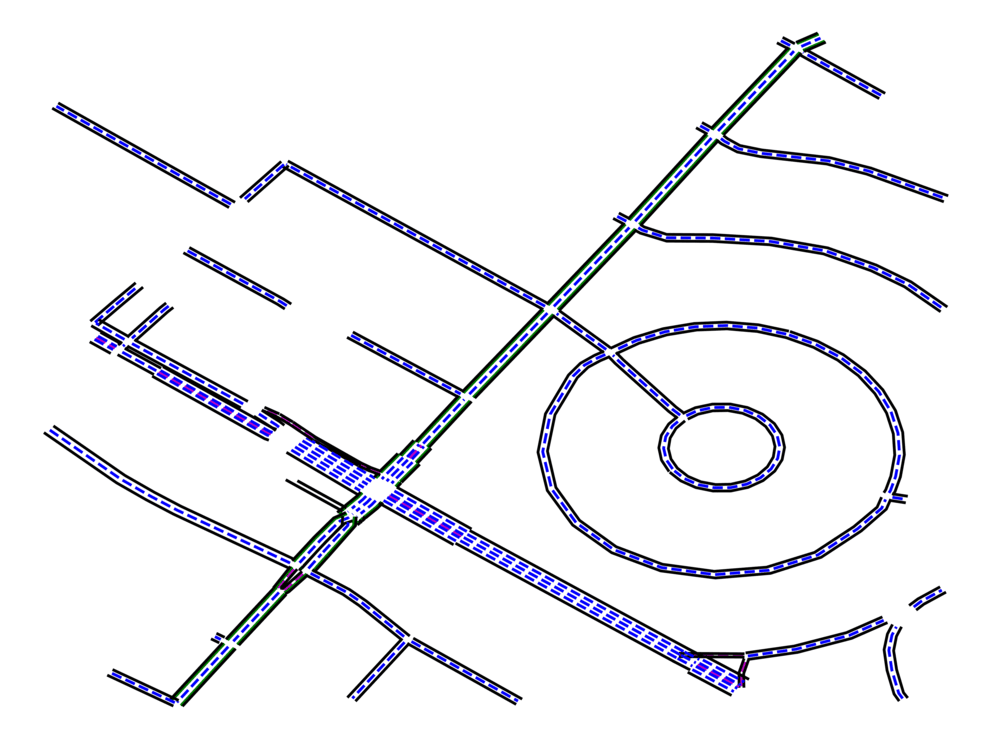}
  \caption{GPT-4o OSG $P2$}
\end{subfigure}
\begin{subfigure}{.49\linewidth}
  \centering
  \includegraphics[width=.95\linewidth]{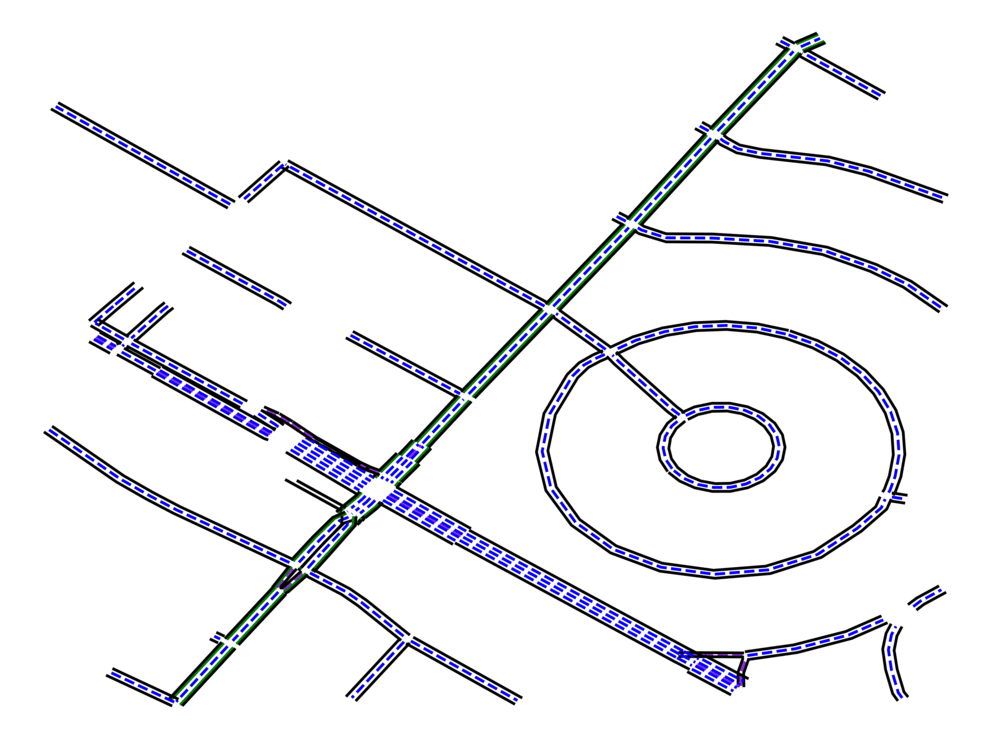}
  \caption{Llama IG}
\end{subfigure}
\begin{subfigure}{.45\linewidth}
  \centering
  \includegraphics[width=.95\linewidth]{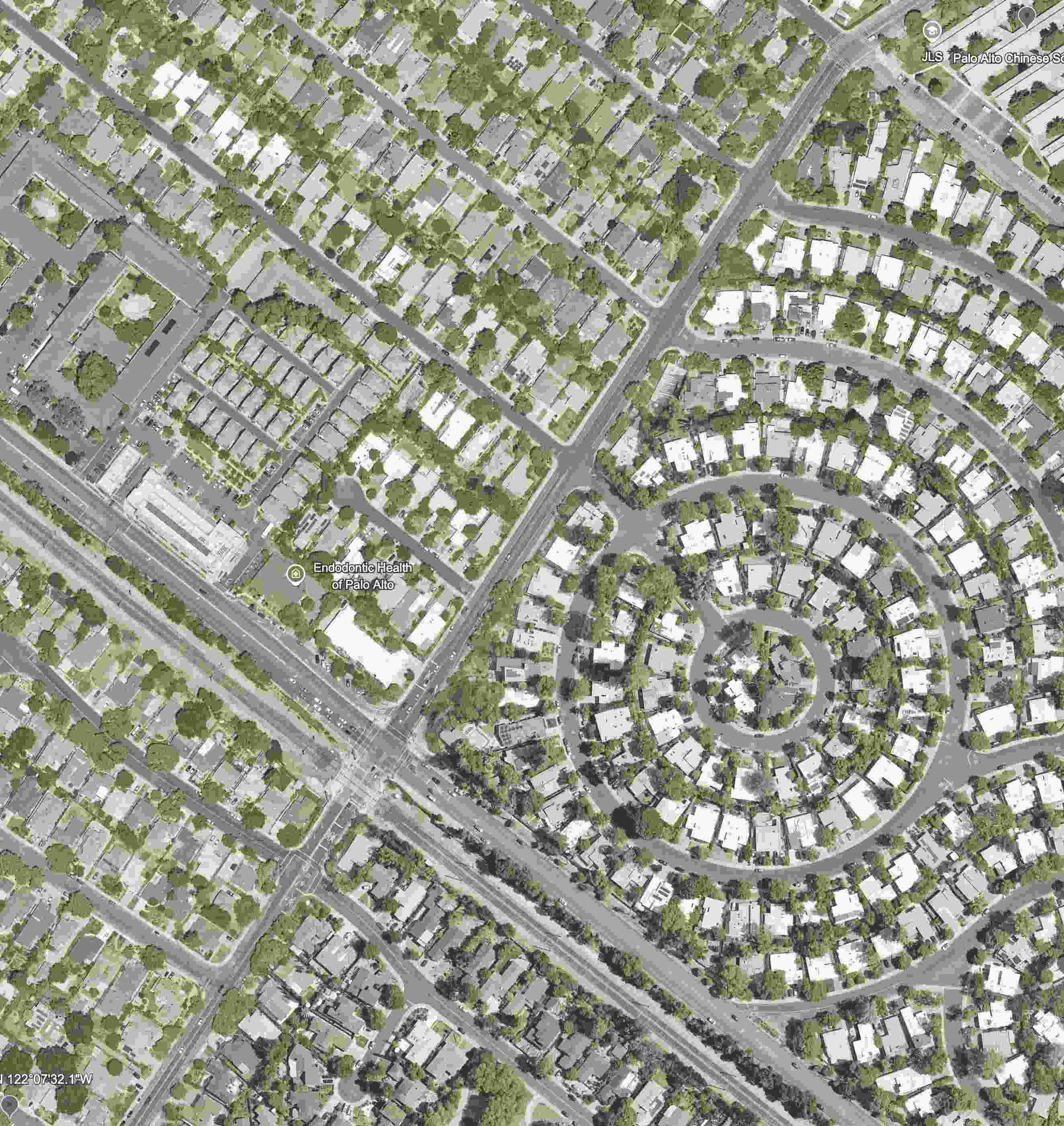}
  \caption{Satellite Image}
\end{subfigure}

\caption{Qualitative comparison for an Argoverse 2 sample from Palo Alto}
\label{fig:av2_qual_cali}
\end{figure}

\begin{figure}[]
\centering
\begin{subfigure}{.5\linewidth}
  \centering
  \includegraphics[width=.95\linewidth]{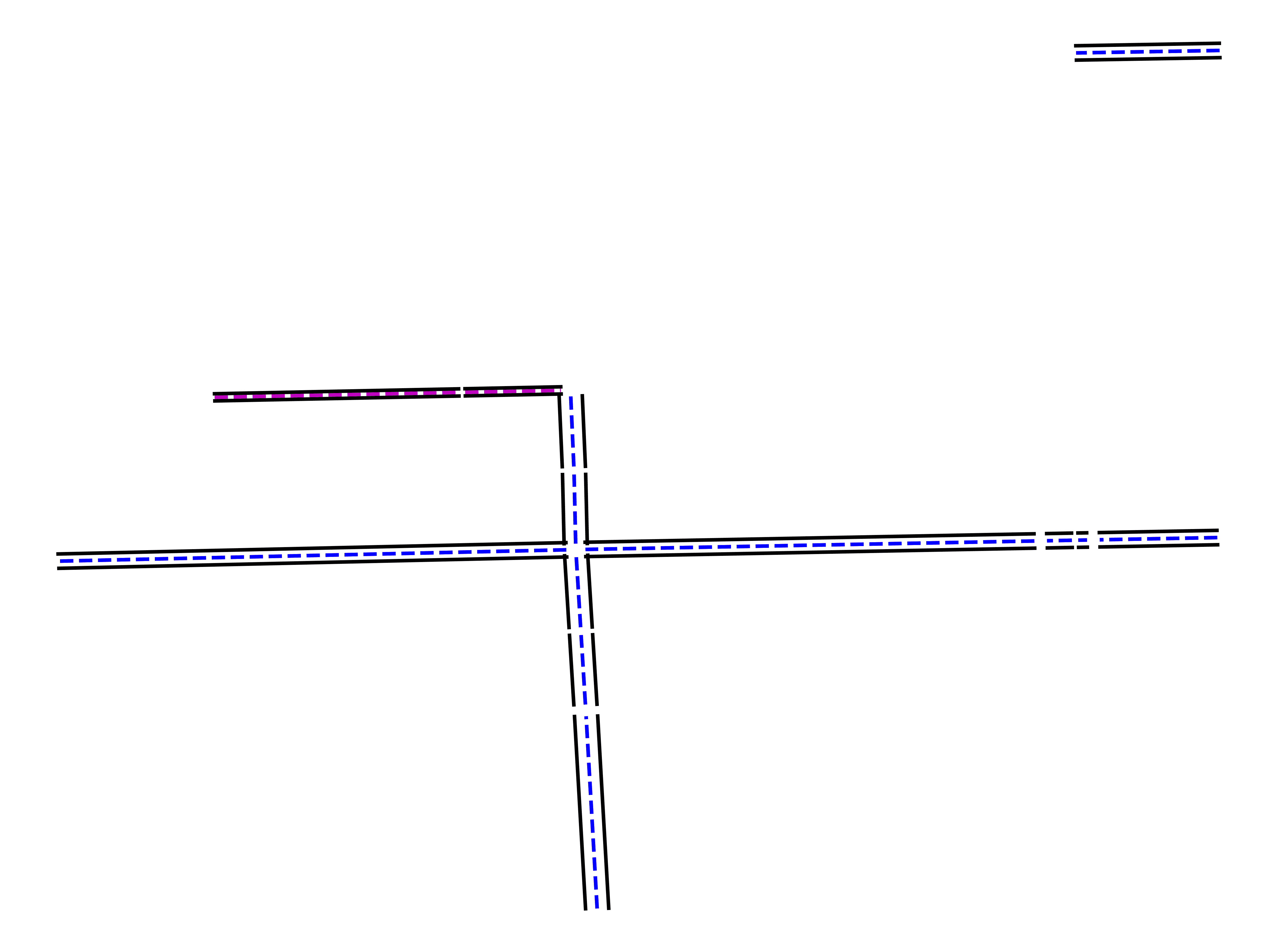}
  \caption{GPT-4o OSG $P1$}
\end{subfigure}%
\begin{subfigure}{.5\linewidth}
  \centering
  \includegraphics[width=.95\linewidth]{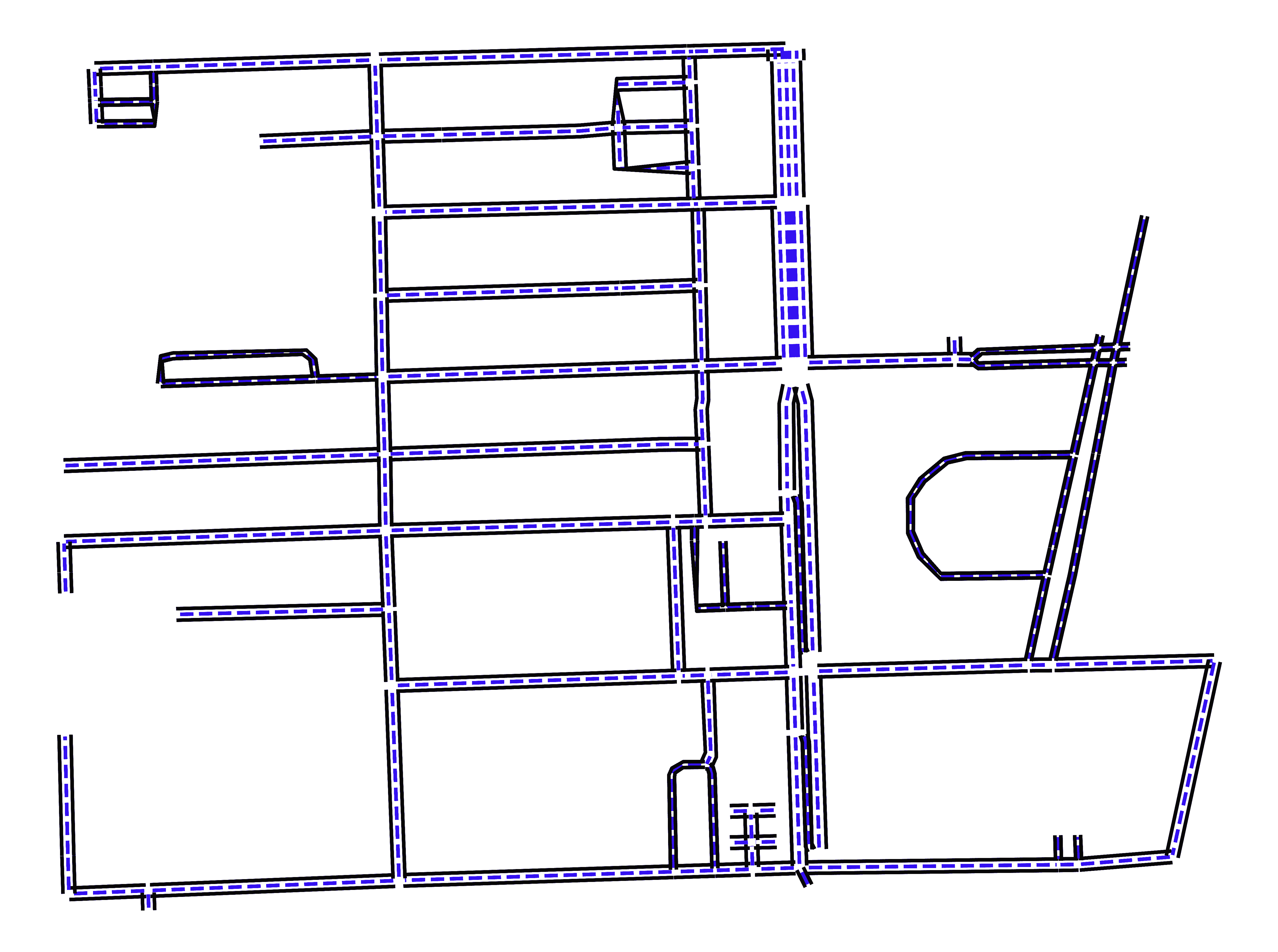}
  \caption{GPT-4o OSG $P2$}
\end{subfigure}
\begin{subfigure}{.5\linewidth}
  \centering
  \includegraphics[width=.95\linewidth]{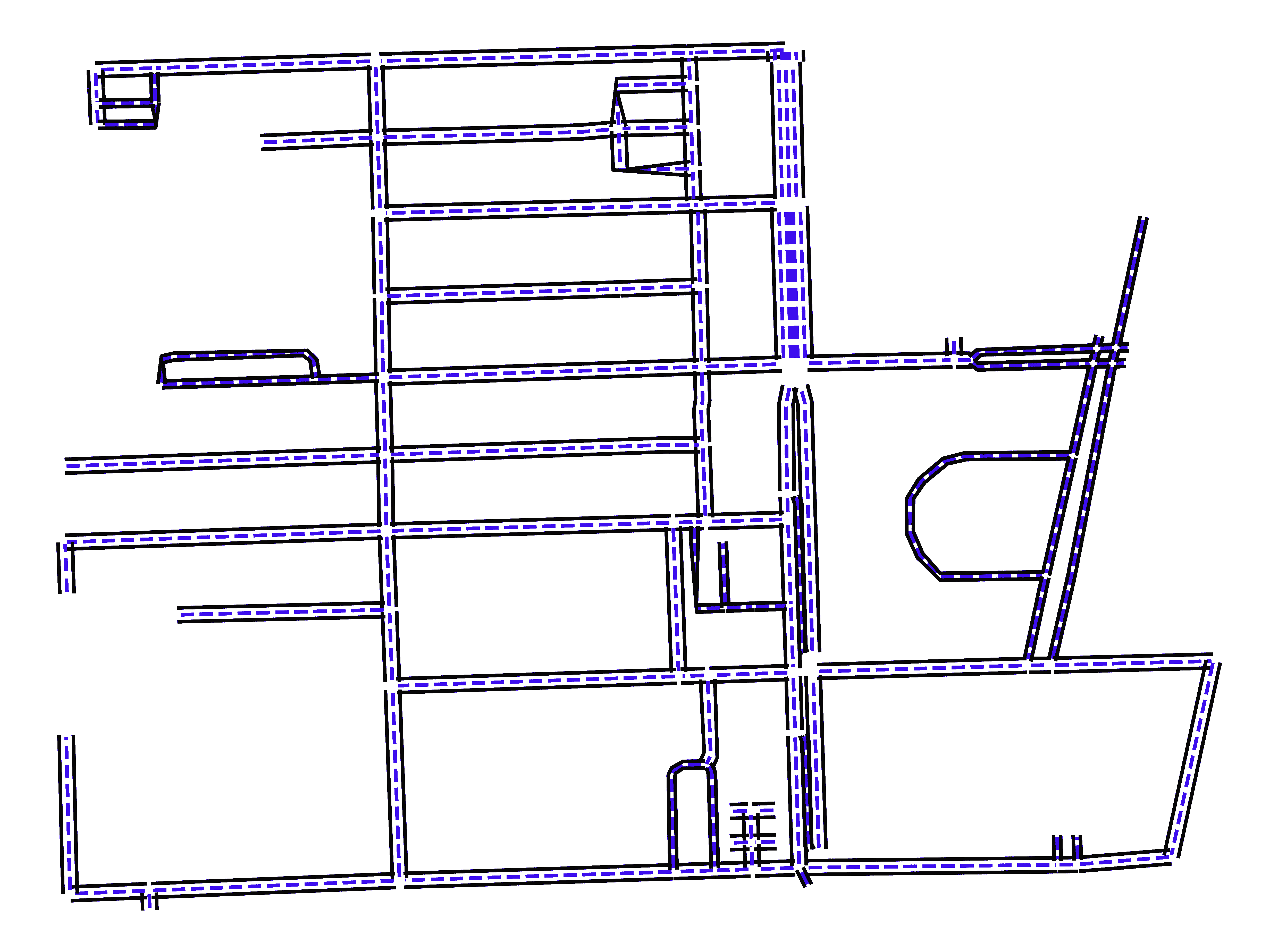}
  \caption{Llama IG}
\end{subfigure}
\begin{subfigure}{.45\linewidth}
  \centering
  \includegraphics[width=.95\linewidth]{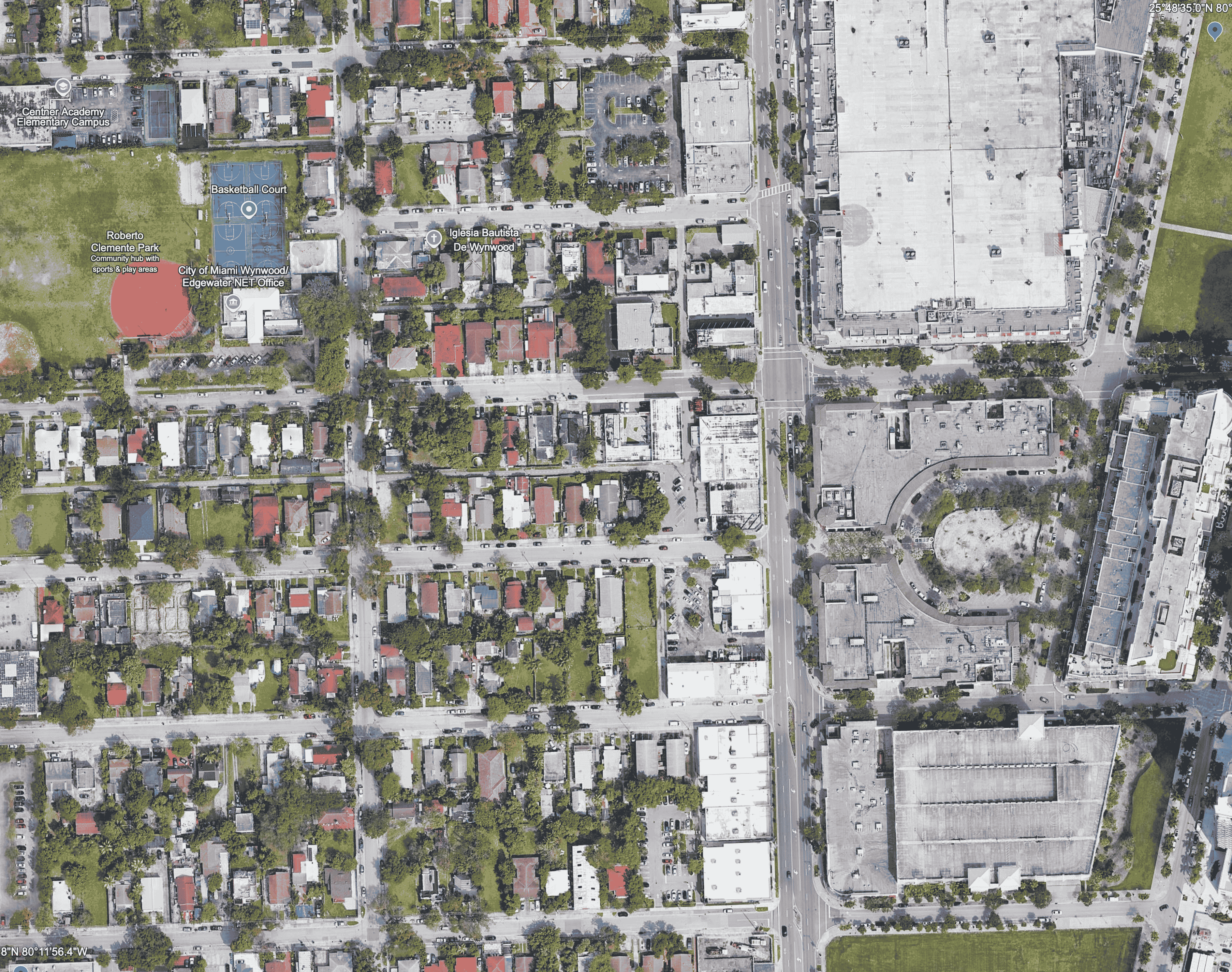}
  \caption{Satellite Image}
\end{subfigure}

\caption{Qualitative comparison for an argoverse 2 sample from Miami}
\label{fig:av2_qual_miami}
\end{figure}

\subsubsection{Qualitative result} To further illustrate the differences between models and prompt variations, we present qualitative results as well as satellite images~\cite{googleearth} in Fig. \ref{fig:av2_qual_cali} and Fig. \ref{fig:av2_qual_miami}. These examples demonstrate how changes in prompt phrasing influence the outputs of both Llama and GPT-4o. As observed in the quantitative analysis, small modifications in prompt wording can lead to significant differences in map generation. For instance, in both Miami and Palo Alto, using prompt $P1$, the model fails to extract all road segments, whereas $P2$ results in a more complete output. 

Additionally, increasing the amount of contextual information in the prompt sometimes leads to a loss in output quality. We hypothesize that the additional input distracts the LLM, making it less effective in generating accurate lane predictions. Furthermore, Llama tends to produce wider lane estimates compared to GPT-4o. We believe this is due to Llama’s smaller model size, which may affect its ability to precisely interpret road design specifications.

Overall, our results highlight the crucial role of both prompt engineering and model selection in improving SD map enhancement. By refining prompt strategies and optimizing iterative processing, we can achieve more reliable and detailed lane structures.



\subsection{Results from Japan} 
\begin{figure}[]
\centering
\begin{subfigure}{.95\linewidth}
  \centering
  \includegraphics[width=.95\linewidth]{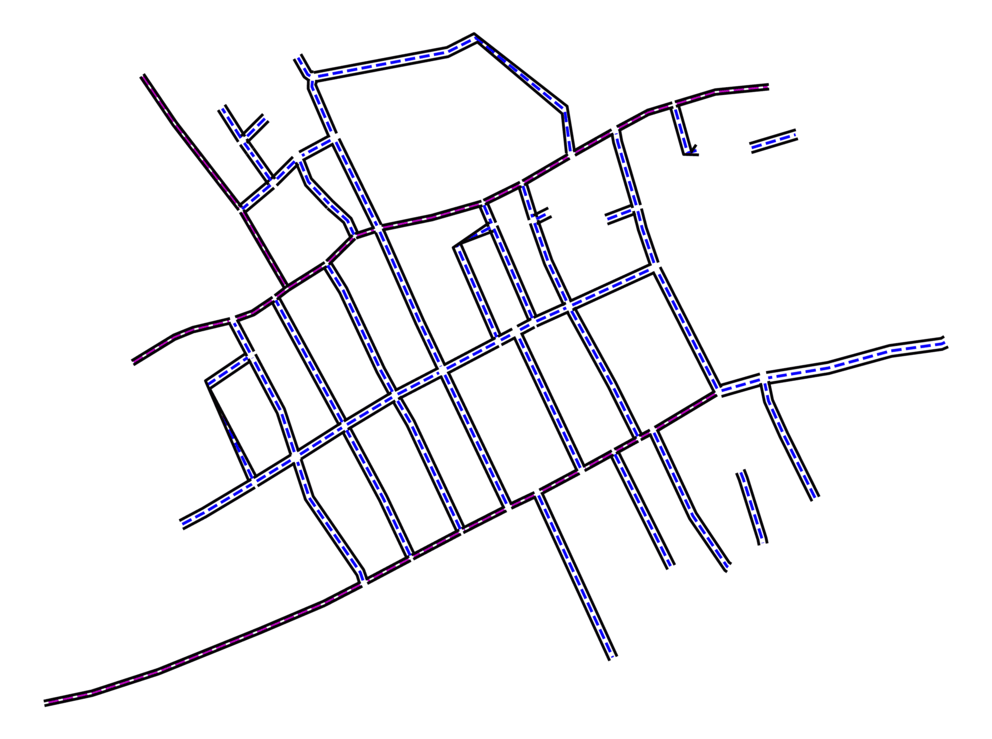}
  \caption{Llama \autoreg}
\end{subfigure}%
\vspace{1.5mm}
\begin{subfigure}{.95\linewidth}
  \centering
  \includegraphics[width=.95\linewidth]{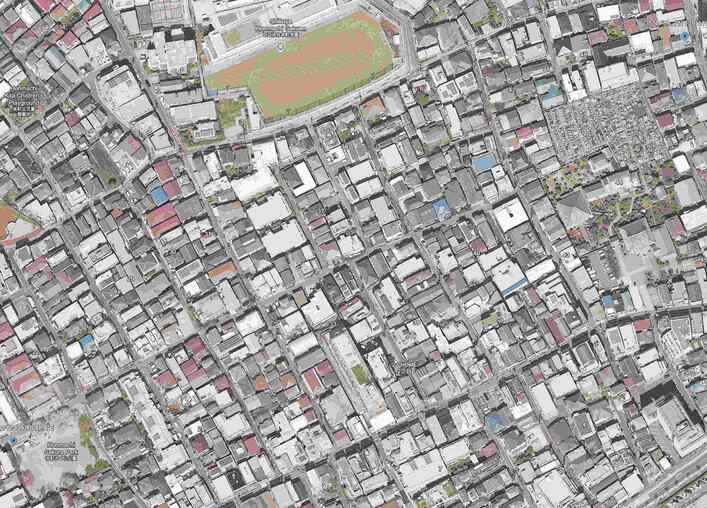}
  \caption{Satellite Image}
\end{subfigure}
\caption{A qualitative example in Japan to demonstrate generalization capability}
\label{fig:japan_results}
\end{figure}

In this section, we demonstrate the generalization capability of \our\ by presenting results from Japan (Fig. \ref{fig:japan_results}). We use an English-translated version of the Japan Road Law~\cite{JapanManual1} as our reference road manual to adapt our approach to a different country. By leveraging this official document, \our\ successfully generates road maps that align with local road design standards, showcasing its ability to adapt across regions.

This result highlights the potential of \our\ to be applied beyond its initial designed region without requiring sensor data or manual annotations. The ability to incorporate diverse road regulations from different countries suggests that our approach can scale efficiently and provide reliable enhanced prior knowledge of roads in various geographic locations.

\section{Limitation and Future Work}
While our method demonstrates strong performance and generalization, it has several limitations. First, the accuracy of SD maps is not improved, as no sensor data is used. Second, the performance of \our\ depends heavily on the quality and completeness of the road manuals. If the manuals lack necessary details, such as specific lane dimensions or configurations, the output may be limited. In cases where OSM does not explicitly show certain road features—such as curved segments or uncommon layouts—\our\ cannot recover that missing information. Additionally, the reliability of our approach is influenced by the LLM's output, which can occasionally be inconsistent or incorrect.

For future work, we plan to improve intersection handling and explore the use of live official sources, such as Caltrans project data~\cite{caltrain}, to improve accuracy and ensure the maps reflect recent changes. We also aim to extend \our\ as a prior for downstream tasks like trajectory prediction and online HD mapping.

\section{Conclusion and Future Work}
In this work, \our\ presents a novel map representation without needing
sensor data. By leveraging LLMs and RAG, we utilize official documents to enhance existing SD maps with lane-level details,
providing valuable context for downstream tasks. \our\ demonstrates
generalizability across OSM data from different states and other countries.
 Overall, we see significant
potential for LLMs in advancing mapping technologies, especially when used as a
for generating strong prior for autonomous driving applications.


\bibliographystyle{unsrt}
\bibliography{ref}

\end{document}

%% file: commands.tex
\newcommand{\red}[1]{{\color{red} #1}}
\newcommand{\our}[0]{SD++}
\newcommand{\oneshot}[0]{OSG}
\newcommand{\increm}[0]{IG}
\newcommand{\autoreg}[0]{IG+Context}